\documentclass[runningheads]{llncs}
\usepackage[T1]{fontenc}
\usepackage{graphicx}
\usepackage{url}
\usepackage[english]{babel}

\usepackage[letterpaper,top=2cm,bottom=2cm,left=1.3cm,right=1.0cm,marginparwidth=2cm]{geometry}

\usepackage{amsmath}
\usepackage{graphicx}
\usepackage[colorlinks=true, allcolors=blue]{hyperref}
\usepackage{colortbl}  
\usepackage{multicol}

\usepackage{framed,multirow}
\usepackage{booktabs}
\usepackage{amssymb}
\usepackage{latexsym}
\usepackage{amsmath}
\usepackage{array}
\usepackage{url}
\usepackage{xcolor}
\usepackage{colortbl}  

\definecolor{mygreen}{RGB}{0,128,0}
\definecolor{myblue}{RGB}{0,0,255}
\definecolor{mybrown}{RGB}{150,75,0}

\begin{document}

\title{CLASS-M: Adaptive stain separation-based contrastive learning with pseudo-labeling for histopathological image classification}
\titlerunning{CLASS-M: Adaptive stain separation-based contrastive learning with pseudo-labeling}

\author{
    \upshape{Bodong Zhang}\textsuperscript{1,2 [0000-0001-9815-0303]} 
    \quad {Hamid Manoochehri}\textsuperscript{1,2 [0009-0005-0478-7925]} 
    \quad {Man Minh Ho}\textsuperscript{2 [0000-0002-7178-8873]} 
    \quad {Fahimeh Fooladgar}\textsuperscript{3 [0000-0002-7859-0456]}
    \quad {Yosep Chong}\textsuperscript{4 [0000-0001-8615-3064]} 
    \quad {Beatrice S. Knudsen}\textsuperscript{5 [0000-0002-7589-7591]} 
    \quad {Deepika Sirohi}\textsuperscript{5 [0000-0002-0848-4172]} 
    \quad {Tolga Tasdizen}\textsuperscript{1,2 [0000-0001-6574-0366]}\\\vspace{5pt}
    {\fontsize{10.5pt}{12pt}\selectfont \textsuperscript{1}Department of Electrical and Computer Engineering, University of Utah, Salt Lake City, UT, USA\\ \textsuperscript{2}Scientific Computing and Imaging Institute, University of Utah, Salt Lake City, UT, USA\\ \textsuperscript{3}Department of Electrical and Computer Engineering, University of British Columbia, Vancouver, BC, Canada
    \\ \textsuperscript{4}Department of Hospital Pathology, College of Medicine, The Catholic University of Korea, Seoul, South Korea
    \\ \textsuperscript{5}Department of Pathology, University of Utah, Salt Lake City, UT, USA
    \\\upshape\texttt{bodong.zhang@utah.edu}
    }}

\authorrunning{B. Zhang et al.}
\institute{}

\maketitle

\begin{abstract}
Histopathological image classification is an important task in medical image analysis. Recent approaches generally rely on weakly supervised learning due to the ease of acquiring case-level labels from pathology reports. However, patch-level classification is preferable in applications where only a limited number of cases are available or when local prediction accuracy is critical. On the other hand, acquiring extensive datasets with localized labels for training is not feasible. In this paper, we propose a semi-supervised patch-level histopathological image classification model, named CLASS-M, that does not require extensively labeled datasets. CLASS-M is formed by two main parts: a contrastive learning module that uses separated Hematoxylin and Eosin images generated through an adaptive stain separation process, and a module with pseudo-labels using MixUp. We compare our model with other state-of-the-art models on two clear cell renal cell carcinoma datasets. We demonstrate that our CLASS-M model has the best performance on both datasets. Our code is available at \url{github.com/BzhangURU/Paper_CLASS-M/tree/main}
\keywords{Contrastive learning\and Stain separation\and pseudo-labeling\and Semi-supervised learning\and Digital histopathological images}
\end{abstract}
\begin{multicols}{2}
\section{Introduction}
Digital histopathological image analysis plays a crucial role in disease diagnosis and treatment optimization. The most commonly used histopathological images are Hematoxylin \& Eosin (H\&E) stained whole slide images (WSIs) that help the differentiation of various tissue sample features ~\cite{Fischer2008-za}. H\&E staining involves the use of two dyes, Hematoxylin (H) and Eosin (E), which selectively stain different components of the tissue samples. Hematoxylin stains the acidic components of the tissue samples, such as the cell nuclei, and has a blue-purple color, while Eosin stains the basic components of the tissue samples, such as the cytoplasm and extracellular matrix, and exhibits a pink color. In recent years, many efforts have been made towards the automatic analysis of H\&E images.

With the rapid development of computing power and advanced algorithms, deep learning has been widely used in the field of digital histopathological image analysis for automating disease diagnosis and performing auxiliary image analysis ~\cite{chan2020deep,litjens2017survey,razzak2018deep,sarvamangala2022convolutional,wang2022medical}. However, achieving a highly accurate model requires a substantial amount of training labels, demanding significant time and effort from human experts. Various strategies have been proposed to mitigate this challenge. Multiple instance learning (MIL), a specific type of weakly-supervised learning, has become the accepted approach in histopathology because of the easy availability of case-level labels from medical reports. Self-supervised learning is also commonly used to train the backbone feature extractor network. 
Nevertheless, semi-supervised learning, despite its success in computer vision, remains a relatively less explored area in histopathology.  

Another direction worthy of further investigation with deep learning models is kidney cancer. Kidney cancer ranks among the most prevalent cancers globally. It is estimated that around 76,080 new cases were diagnosed with cancers of the kidney and renal pelvis and there are around 13,780 deaths resulting from it in the US in 2021 ~\cite{motzer2022kidney}. 
Clear cell renal cell carcinoma (ccRCC) stands out as the predominant subtype that dominates the kidney cancer cases ~\cite{bian2022novel}. Consequently, further research on cancer detection and classification using ccRCC images holds paramount importance for disease diagnosis and early patient treatment. In this paper, we experiment with two ccRCC datasets. Even though MIL approaches, requiring only slide-level annotations, are widely used in the histopathology field ~\cite{laleh2022benchmarking,li2023vision,qian2022transformer,wang2022weakly}, they require a large number of WSIs for effective training. For instance, the Utah ccRCC dataset we experiment with consists of only 49 WSIs. Furthermore, if multiple classes always co-occur in WSIs, MIL approaches can not differentiate between them. This is another issue with the Utah ccRCC dataset which contains only cancer cases; therefore, normal regions always co-occur with cancer regions in WSIs. Finally, MIL methods typically classify WSIs based on only the most discriminative patches leading to poor patch-level accuracy despite demonstrating good slide-level accuracy. This can be a problem even for larger datasets such as The Cancer Genome Atlas Program (TCGA) ccRCC dataset we experiment with which comprises 420 WSIs.

In our approach, pathologists draw approximate polygons at a low resolution to mark regions in a subset of WSIs and assign labels to those regions. We crop patches from inside these annotated regions to collect labeled samples for performing patch-level classification tasks. To fully utilize the WSIs, we also collect patches from outside the annotated regions and from WSIs that are not annotated to gather unlabeled samples. Consequently, we form a semi-supervised classification task based on the labeled and unlabeled patches. Notably, a patch-level classification model serves as a precise tool to expedite pathologists' identification of cancerous regions within WSIs.

In our work, we propose a new semi-supervised model for histopathological image classification: Contrastive Learning with Adaptive Stain Separation and MixUp (CLASS-M), where CLASS-M can also be understood as a model for CLASSifying Medical images. CLASS-M is an extension of ~\cite{zhang2022stain} presented in MICCAI 2022 Workshop on Medical Image Learning with Limited and Noisy Data with substantial modifications, a more thorough experimental analysis and significant improvements in classification accuracy. A simple and effective novel unsupervised loss is formulated for contrastive learning between adaptively stain-separated Hematoxylin images and Eosin images. Additionally, considering the benefit of pseudo-labeling in state-of-the-art general semi-supervised learning methods ~\cite{sohn2020fixmatch} ~\cite{berthelot2019mixmatch}, pseudo-labeling with MixUp ~\cite{zhang2017mixup} is adopted to provide further regularization via additional samples and pseudo-labels in training. The two methods benefit from different regularization effects, contrastive learning with multiple views vs. data augmentation, and hence can be combined to better utilize the information in both labeled and unlabeled samples and improve classification accuracy. Furthermore, instead of using a globally fixed stain separation matrix in Optical Density space as in ~\cite{zhang2022stain}, a simple and efficient slide-by-slide stain separation based on ~\cite{macenko2009method} is applied to adapt to properties of individual WSIs. We not only perform image augmentations on Hematoxylin images and Eosin images, but also apply augmentations on original RGB images before the stain separation to strengthen the augmentation process. Finally, in this paper, we compare our method to multiple state-of-the-art semi-supervised and self-supervised learning models on the Utah ccRCC dataset and TCGA ccRCC dataset. 
We also conduct ablation studies to carefully analyze the contributions of different parts in our model. 

In conclusion, the main contributions of our paper are as follows:
\begin{itemize}
    \item We propose the CLASS-M model for semi-supervised learning. We apply our novel contrastive learning on Hematoxylin images and Eosin images after performing slide-level stain separation. We also use pseudo-labeling with MixUp to further improve classification accuracy. 
    \item We provide the new Utah ccRCC dataset which has 49 WSIs with patch-level labels, and patch-level labels for 150 WSIs from the TCGA ccRCC dataset. The Utah ccRCC dataset and its annotations will be made available upon transfer agreement. The annotations on TCGA ccRCC dataset are made publicly available.
    \item We test various state-of-the-art semi-supervised learning and self-supervised learning 
    methods on both ccRCC datasets and demonstrate that CLASS-M outperforms other state-of-the-art models.
    \item The code for our paper is available at \url{github.com/BzhangURU/Paper_CLASS-M/tree/main}
    
\end{itemize}

\section{Related work}
\subsection{Weakly-supervised learning}
Weakly-supervised learning ~\cite{kanavati2020weakly,oquab2015object,zhou2018brief} leverages slide or case-level labels to help guide the training process. A typical case of weakly-supervised learning is MIL ~\cite{foulds2010review,herrera2016multiple}. 
Various MIL approaches, such as instance-based and embedding-based methods, have been proposed. The instance-based approach~\cite{hou2016patch,chikontwe2020multiple} involves modeling a classifier at the instance level, then aggregating the predicted instance labels to form bag label predictions. However, due to the noise in instance labels, the performance of this aggregation may be impacted. On the other hand, the embedding-based approach~\cite{hashimoto2020multi,li2021multi} first creates a bag representation from individual instance representations, followed by training a classifier on these bag representations. Research has shown that embedding-based approaches are generally more effective than instance-based methods.
The attention-based MIL method, proposed in ~\cite{ilse2018attention}, suggests that instead of treating all patches within a bag equally, assigning varying importance scores to patches, particularly the more discriminative ones, is more effective. This method involves computing attention scores from instance representations to reflect the importance of respective patches and forming the bag representation through a weighted average of these instances.  ~\cite{xiong2023diagnose} proposed a hierarchical attention-guided MIL that effectively identifies important areas at different scales in WSIs. This method combines several attention techniques to form a comprehensive group representation. Furthermore, the recent Self-ViT-MIL approach by ~\cite{gul2022histopathological}, which combines self-supervised learning with Vision Transformers (ViTs) ~\cite{dosovitskiy2020image} and MIL, has shown promising results, even in comparison to fully-supervised methods.

In digital histopathological image applications, MIL can be applied when only whole slide image-level labels are provided, but the specific regions that contribute to the labels are not given. Despite the convenience of annotation work, weakly-supervised learning can't deal with small datasets so well, as it needs an adequate amount of slides for learning. Additionally, weakly-supervised learning struggles with multiple classes which almost always co-exist in slides or a certain class that lacks positive or negative labels in datasets. For example, in our Utah ccRCC dataset, cancer is consistently present in all WSIs, resulting in a lack of negative labels for the cancer category in the context of MIL. 

\subsection{Self-supervised learning}
Self-supervised learning (SSL) ~\cite{jaiswal2020survey,krishnan2022self} extracts useful patterns from data itself without explicit labels provided by humans. Self-supervised learning allows models to first pre-train on a large unlabeled dataset, where effective feature representations can be learned. Examples of pre-training tasks include image inpainting, predicting rotations, and colorizing images. Then the final layers of the models can be optimized in specific downstream tasks, reducing the need for extensively labeled data. SimCLR ~\cite{chen2020simple,chen2020big} presents a seminal contrastive learning to train feature representation, where augmented views of an image (positive samples) are minimized within that image while being maximized against augmented views of other images (negative samples). MoCo ~\cite{chen2020improved,chen2021empirical,chen2020mocov2,he2019moco,he2020momentum,he2022masked} strengthens SSL with a dynamic dictionary, tapping into past batches for more negative samples, and a slowly updating momentum encoder for stable global representations. In contrast, Barlow Twins ~\cite{zbontar2021barlow} omits negative samples, focusing on refining features by measuring cross-correlation between positive samples. BYOL ~\cite{grill2020bootstrap} and DINO ~\cite{caron2021dino} utilize a teacher-student setup, where the student refines its understanding by predicting the teacher's representation, while the teacher network is updated through the slow-moving average of student’s parameters. Clustering-based methods like Deep Clustering ~\cite{caron2018deepclustering} and SwAV ~\cite{caron2020swav} define pseudo-labels, sorting images into clusters, yielding great performance on downstream tasks. 

In histopathology image analysis, self-supervised learning can be used to provide the backbone feature extraction module for MIL and patch-level classification models. Mingu et al. ~\cite{kang2023benchmarking} benchmarked SSL methods on pathology datasets, showcasing consistent enhancements in histopathology tasks. 
CS-CO ~\cite{yang2022cs} also utilizes separate encoders for Hematoxylin images and Eosin images. Instead of contrastive loss between Hematoxylin images and Eosin images introduced in our work, CS-CO calculates a contrastive loss between different augmentations. In addition, CS-CO is a self-supervised learning method that requires encoders to generate a visual representation containing enough information to recover images in cross-stain prediction, while our semi-supervised CLASS-M only pursues shared latent features containing enough information to perform classification tasks. 
While self-supervised learning benefits from unlabeled data, semi-supervised learning has the advantage that it allows both unlabeled data and labeled data to be trained at the same time and use the knowledge learned from labeled data to better utilize unlabeled data. 

\begin{figure*}
    \centering
    \includegraphics[width=0.8\textwidth]{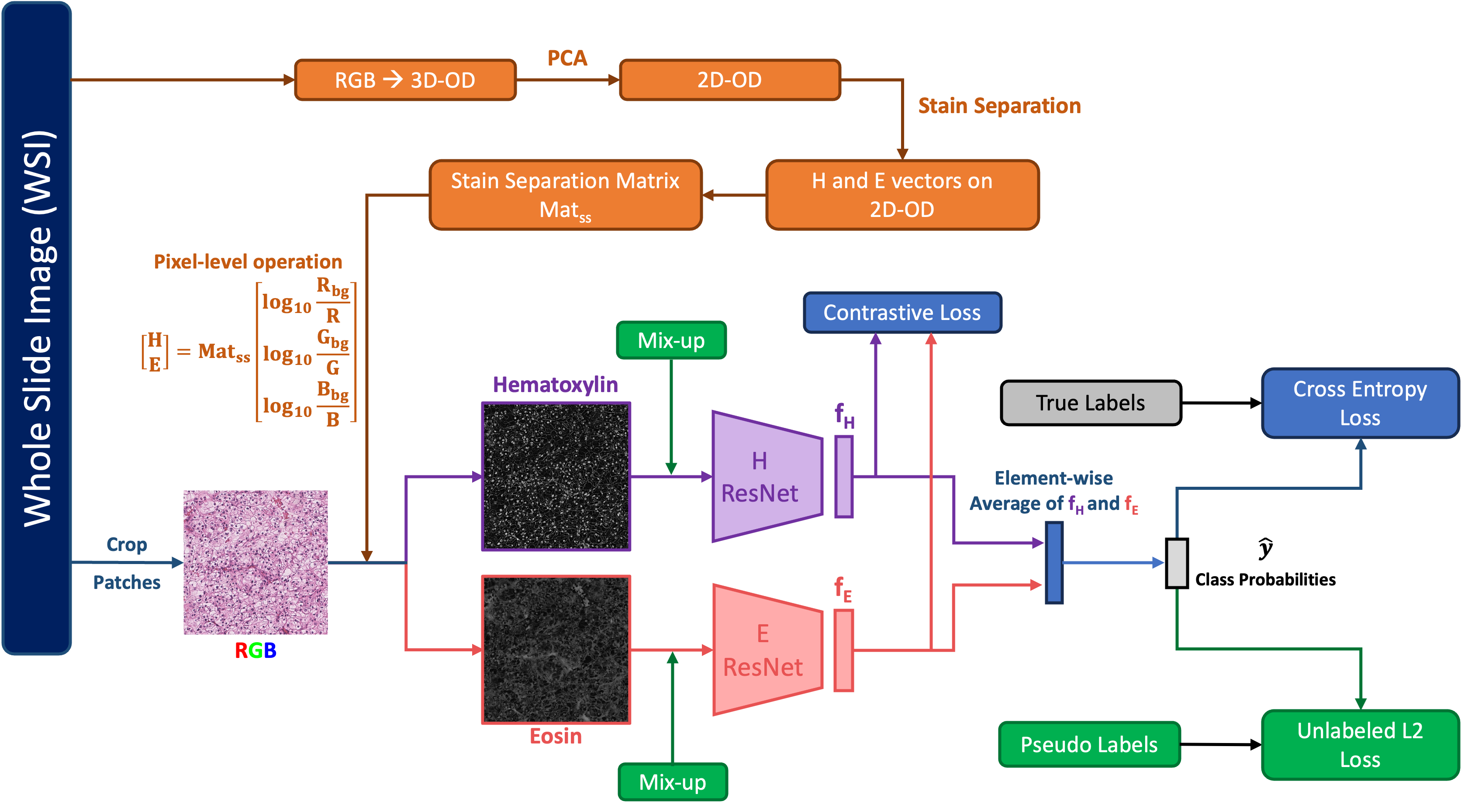}
    \caption{The block diagram of our Contrastive Learning with Adaptive Stain Separation and MixUp (CLASS-M) for semi-supervised histopathological image classification. Orange part shows adaptive stain separation, where OD means Optical Density space. Green part shows mixup on both labeled and unlabeled samples. }
    \label{fig:flowchart}
\end{figure*}

\subsection{Semi-supervised learning}
Semi-supervised learning ~\cite{chartsias2020disentangle,ouali2020overview,van2020survey} involves training a model simultaneously with both labeled and unlabeled data to enhance training outcomes, which reduces reliance on a limited number of annotated samples. Semi-supervised learning finds applications when obtaining fully labeled data is prohibitively expensive or impractical. For digital histopathological image analysis, getting a large amount of labeled data requires great effort from well-trained experts, which is expensive and time-consuming. However, it is much easier to acquire unlabeled data such as unannotated WSIs. One type of semi-supervised learning is consistency regularization which seeks agreement among the model predictions for the same input with different views, augmentations or epochs. For example, temporal ensembling ~\cite{laine2016temporal} aims to reach a consensus in the prediction of labels between current epochs and previous epochs. ~\cite{sajjadi2016regularization} tries to make classifier's prediction consistent across various transformations. More recent consistency regularization methods, such as FixMatch ~\cite{sohn2020fixmatch} and MixMatch ~\cite{berthelot2019mixmatch}, have been proposed. FixMatch assigns pseudo-labels to unlabeled samples when the prediction confidence is high with weak augmentation of input data. Then these pseudo-labels are used for training with strong augmentation. MixMatch is another consistency regularization method where pairs of labeled/unlabeled input data are linearly combined to create a "mixup" of data. The label of newly created data is a weighted average of the original labels followed by a sharpening process. Inspired by self-supervised learning, contrastive learning has also been integrated in semi-supervised learning ~\cite{alonso2021semi,chaitanya2023local,singh2021semi,zhang2022stain}. Instead of seeking agreements in label predictions, contrastive learning focuses on the representation of positive and negative pairs. Positive pairs originate from the same inputs with different views or augmentations, while negative pairs involve different inputs. The method encourages a shorter distance between positive pairs and a longer distance between negative pairs in the feature representation space. In this paper, we propose a novel histopathology-specific contrastive learning based semi-supervised classification model.

\section{Methods}
\subsection{An overview of CLASS-M}
Figure \ref{fig:flowchart} shows the workflow of CLASS-M. As shown in the orange part of Figure \ref{fig:flowchart}, slide-level stain separation is applied for generating Hematoxylin images and Eosin images from original RGB images. The original RGB images, Hematoxylin images and Eosin images are all augmented during training to improve the robustness of CLASS-M model. The model takes Hematoxylin images and Eosin images as input pairs to form different views of input data. The purple and pink boxes in Figure \ref{fig:flowchart} illustrate the H and E ResNet ~\cite{he2016deep} encoders that have the same architecture but separate parameters to generate latent features $f_H$ and $f_E$. A contrastive loss is proposed for shared latent feature space between H channel and E channel to increase the similarity between $f_H$ and $f_E$. We take the average of features $f_H$ and $f_E$ and pass it to a linear+softmax layer to predict the labels. For labeled samples in training, the cross-entropy loss is introduced to measure the difference between predictions and true labels. Moreover, a challenge in semi-supervised learning in general is the limited number of labeled samples. Pseudo-labeling on unlabeled samples has been widely used in many state-of-the-art semi-supervised learning methods ~\cite{berthelot2019mixmatch,sohn2020fixmatch,zhang2021flexmatch}. In MixMatch ~\cite{berthelot2019mixmatch}, after pseudo-labeling on original samples, the model further uses MixUp ~\cite{zhang2017mixup} to introduce virtual samples by linear interpolation of two random samples. Inspired by those ideas and to fully utilize unlabeled samples, we first provide pseudo-labels to unlabeled samples, then add MixUp on both labeled and unlabeled samples to create virtual samples. The labels of the generated samples after MixUp are set to the weighted averages of original labels/pseudo-labels with a sharpening process. The green part in Figure \ref{fig:flowchart} shows the workflow of this process for pseudo-labeling. We introduce different parts of CLASS-M model in detail in the following subsections. 

\subsection{Adaptive stain separation}
Stain separation is introduced to separate different types of stains present in histological images ~\cite{Perez-Bueno2022-iw,Vasiljevic2021-xd,Zhang2022-ws-notB}, such as separating Hematoxylin and Eosin (H\&E) images into Hematoxylin images and Eosin images. The process has many challenges due to the variation of stains caused by different manufacturers, storage conditions and staining procedures. In our model, a simple and unsupervised method for adaptive stain separation ~\cite{macenko2009method} is applied. The main idea is to first project all pixels from RGB space to Optical Density (OD) space, where stain components are formed by linear combinations. Inside the OD space, the appropriate Hematoxylin and Eosin vectors are calculated. Then all the RGB pixels can be projected to the Hematoxyling portion and Eosin portion to form Hematoxylin images and Eosin images. The images are further normalized to mitigate the effect of variations on the brightness and dosage of stains. A more detailed description of the algorithm can be found in Appendix A.

\begin{figure*}
    \centering
    \includegraphics[scale=.25]{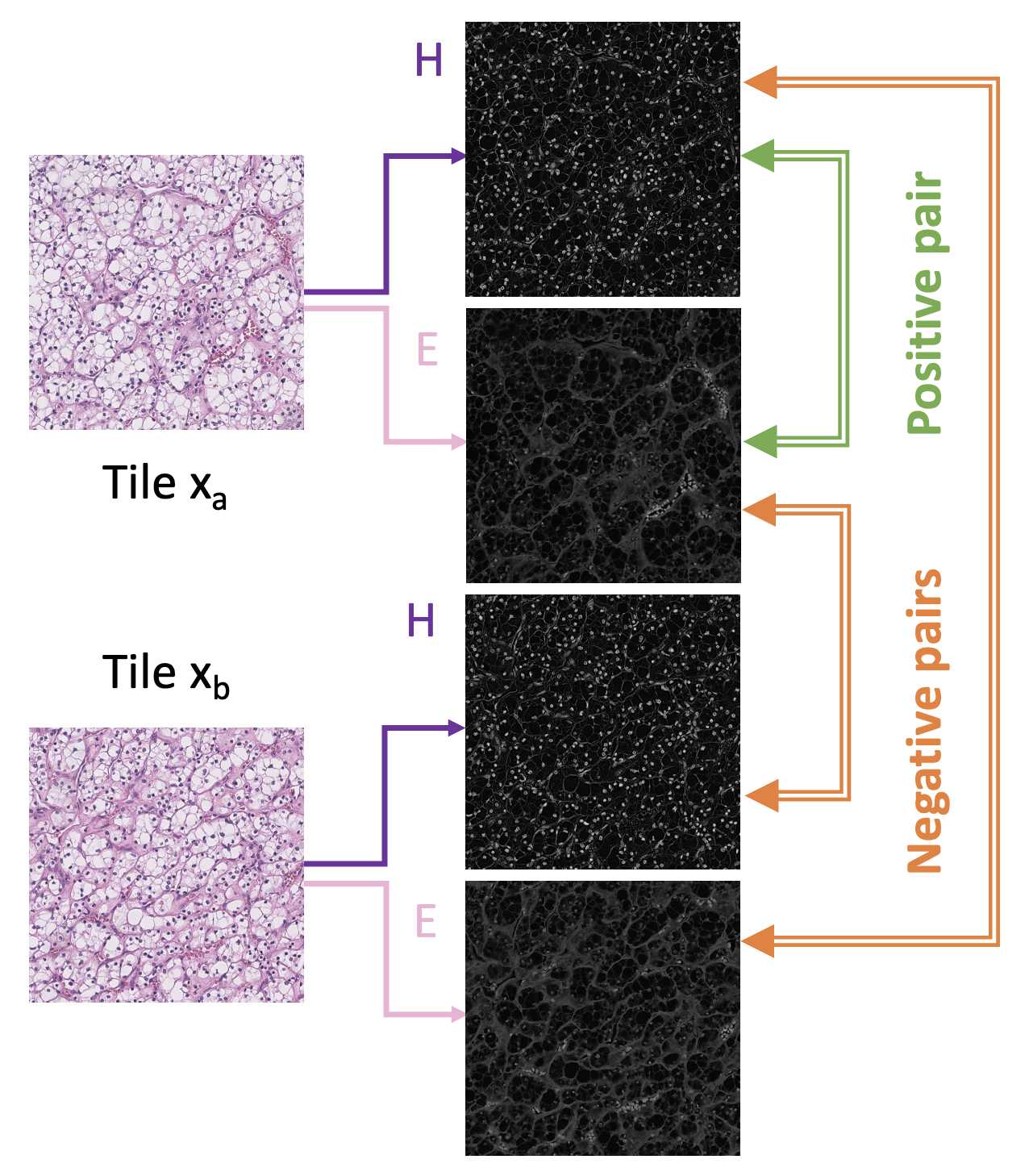}
    \caption{Contrastive learning on Hematoxyling images and Eosin images}
    \label{fig:Cont_loss}
\end{figure*}

\subsection{Augmentations on original RGB, Hematoxylin and Eosin images}
We perform color-jittering augmentations on brightness, contrast and saturation of original images, followed by image augmentations on Hematoxylin images and Eosin images, including random crops, random rotations and random flips. Independent jittering of brightness on Hematoxylin images and Eosin images is also applied. The extra augmentations on the original RGB images provide increased diversity in the training samples and help learn more general features. 

\subsection{Contrastive learning}
 Inputs from two different views are usually adopted in co-training. The general presumptions of co-training are that each input view carries sufficient information for the task and the views should be highly independent of each other to achieve good results. Our preliminary work ~\cite{zhang2022stain} has demonstrated that Hematoxylin and Eosin channels fulfill these requirements in contrastive co-training.

Let $f_H(x)$ and $f_E(x)$ denote the output features after the two ResNet models that take sample x as input. Then the contrastive loss function on sample $x_i$ is written as 
\begin{equation}
\begin{split}
&{\cal L}_{ct}(x_i)= \\
&\max\left(
\parallel f_H(x_i) - f_E(x_i)\parallel_2 - \parallel f_H(x_i) - f_E(x_k)\parallel_2 + m
,0\right)
\label{eqn:triplet}
\end{split}
\end{equation}
where $x_k$ represents another random sample, $\parallel a\parallel_2$ denotes the L2 norm of vector $a$, and $m$ serves as a margin hyperparameter. This contrastive loss term forms a positive pair whose $f_H(x)$ and $f_E(x)$ originate from the same sample and a negative pair whose $f_H(x)$ and $f_E(x)$ come from different samples. Figure \ref{fig:Cont_loss} shows an example of how the positive pairs and the negative pairs are formed. 

During training, the H and E features from the same sample are pulled closer in the shared latent feature space, while the features from different samples are pushed further away. 

\subsection{MixUp}
MixUp ~\cite{zhang2017mixup} is applied on both labeled and unlabeled samples. First, 
we assign pseudo-labels to unlabeled samples. Assuming $x_i$ is an original unlabeled sample from unlabeled training set $U$, we augment it $K$ times and get the average of softmax predictions from the model. The result is sharpened to lower the entropy of label distribution using
\begin{equation}
    \bar{y_i}=\frac{1}{K}\sum_{1\leq k\leq K} P_{model}(x_{i,k}), x_i \in U
    \label{eqn:avg_pred}
\end{equation}
\begin{equation}
    y_i=sharpen(\bar{y_i},T), \text{where}\; sharpen(y,T)_{(c)}=\frac{{y_{(c)}}^{\frac{1}{T}}}{\sum_{j=1}^{C}{y_{(j)}}^{\frac{1}{T}}}
    \label{eqn:sharpen1}
\end{equation}
where $x_{i,k}$ iz the $k$-th augmentation of sample $x_i$, $T$ is a temperature hyperparameter to control the sharpness, $y_{(c)}$ is the $c$-th element in one-hot label encoding $y$, $C$ is the  number of classes, and $y_i$ is the virtual one-hot label encoding of sample $x_i$. 

We follow the standard procedure of MixUp to mix samples. Let $x_i$ and $x_j$ be two random samples from labeled training set L or unlabeled training set U, whose one-hot label encodings are $y_i$ and $y_j$. Let $\lambda$ denote a random number from Beta distribution $Beta(\alpha,\alpha)$, and set $\lambda^\prime=max(\lambda,1-\lambda)$. Then the new mixed sample and its label are as follows:
\begin{equation}
    x^\prime=\lambda^{\prime}x_i+(1-\lambda^{\prime})x_j
    \label{eqn:mixup_x}
\end{equation}
\begin{equation}
    y^\prime=\lambda^{\prime}y_i+(1-\lambda^{\prime})y_j
    \label{eqn:mixup_x}
\end{equation}
After MixUp, we define $x^{\prime}$ as labeled sample if $x_i$ is from labeled set to form new labeled set $L^{\prime}$, the remaining samples form new unlabeled set $U^{\prime}$ with virtual labels. Since $\lambda^{\prime}$ is random, sample diversity is greatly increased. 

\subsection{Loss function}
The total loss is formed by the summation of losses from label prediction and contrastive learning. Inspired by MixMatch ~\cite{berthelot2019mixmatch}, we use cross-entropy loss in labeled set $L^{\prime}$ and squared $L_2$ loss between predictions and virtual labels in unlabeled set $U^{\prime}$ considering squared $L_2$ loss is less sensitive to incorrect predictions. The total batch loss can be written as 
\begin{equation}
    {\cal L}=\sum_{x_i \in L^{\prime}} \frac{y_i\log \hat{y}_i}{|L^{\prime}|}+{\lambda}_{U^\prime}\sum_{x_i \in U^{\prime}} \frac{\parallel y_i - \hat{y}_i {\parallel_2}^2}{C|U^{\prime}|}
    + {\lambda}_C \sum_{x_i \in L^{\prime}\cup U^{\prime}} {\cal L}_{c.t.}(x_i)
    \label{eqn:loss}
\end{equation}
$y_i$ is one-hot encoding of labels (in $L^\prime$) or virtual labels (in $U^\prime$), $\hat{y}_j$ is the label prediction from the model, C is the number of classes, $|L^\prime|$ and $|U^\prime|$ are the size of labeled set and unlabeled set in a batch. $\lambda_{U^\prime}$ and $\lambda_C$ are hyperparameters that control the weights of squared $L_2$ loss and contrastive learning loss. 

\begin{figure*}
    \centering
    \includegraphics[width=1.0\textwidth]{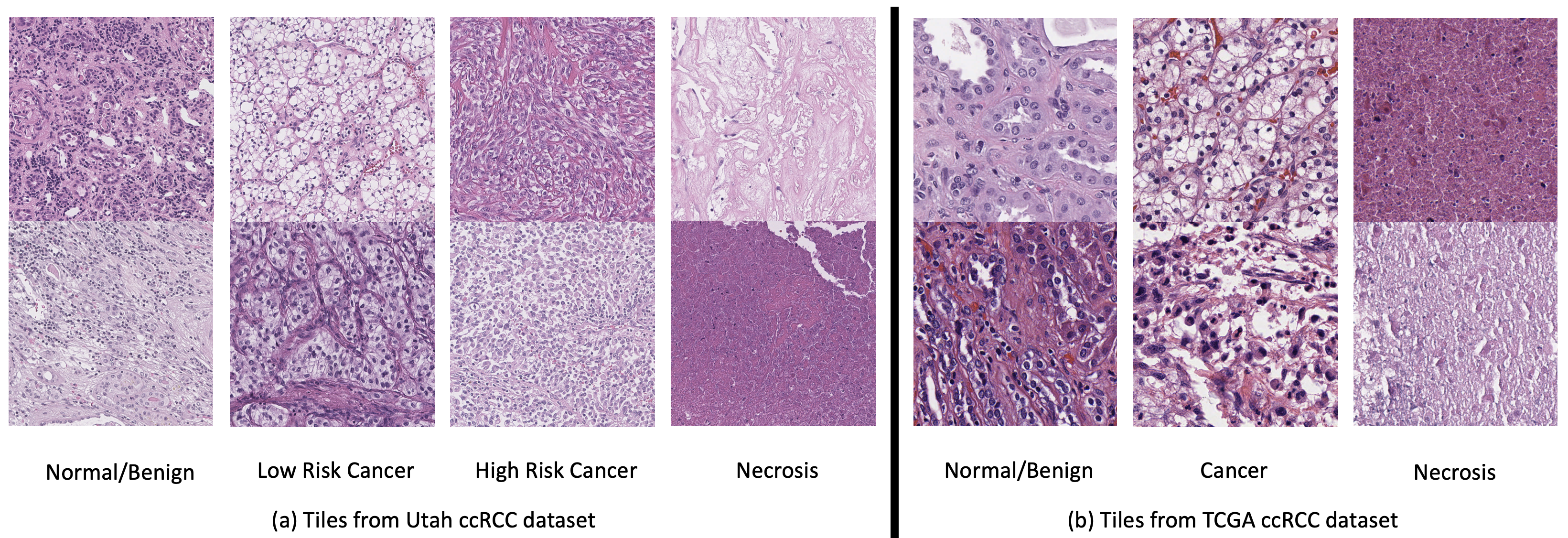}
    \caption{Examples of tiles with the size of 400$\times$400 from (a) Utah ccRCC dataset (in 10X) and (b) TCGA ccRCC dataset (in 20X). }
    \label{fig:institution_TCGA_tiles}
\end{figure*}

\section{Experiments}
\subsection{Datasets}
To evaluate our method, we applied  CLASS-M on the Utah ccRCC dataset and TCGA ccRCC dataset separately, and compared results with other semi-supervised and self-supervised classification methods. 

In the Utah ccRCC dataset, there are 49 WSIs from 49 patients. First, a pathologist drew polygons inside WSIs to mark areas with certain growth pattern labels, which were subsequently verified by another pathologist. We randomly split the WSIs into 32, 10, and 7 WSIs for training, validation and test set. The WSIs were then cropped into 400$\times$400 tiles at 10X resolution with stride set to 200 pixels inside each labeled polygon. The same process was applied to crop tiles outside polygons in 32 training WSIs to collect unlabeled tiles to form a semi-supervised learning task for a 4-class classification. In detail, there are 28497, 2044, 2522, 4115 tiles in the category of Normal/Benign, Low risk cancer, High risk cancer, and Necrosis, respectively, in the labeled training set extracted from polygons in 32 WSIs. Additionally, there are 171,113 unlabeled training tiles, 5472, 416, 334, 2495 tiles respectively for each category in the validation set, and 7263, 598, 389, 924  tiles respectively for each category in the test set. Tiles with predominantly background areas were removed, along with those tainted by ink. Tile examples can be seen in Figure \ref{fig:institution_TCGA_tiles}(a). \footnote{The Utah ccRCC dataset is available through a transfer agreement by contacting the authors.}

For the TCGA ccRCC dataset, we have in total 420 WSIs from 420 patients. 150 of them were labeled by a pathologist through drawing polygons with annotations and verified by a second pathologist. The remaining 270 WSIs were treated unlabeled. If pathologists had disagreements or concerns about labels of polygons, then those annotations were abandoned. The resolution we trained on is 20X. To make it a more challenging task and show the effectiveness of semi-supervised learning, we split the 150 labeled WSIs into 30, 60 and 60 WSIs for training, validation and test set. The tile cropping process was the same as in the Utah ccRCC dataset except for different strides. We chose 200 pixels as stride for labeled training set, but 400 pixels for validation and test set, considering they contain a large number of tiles. We cropped foreground tiles outside polygons in the 30 labeled training WSIs, as well as foreground tiles across the 270 unlabeled WSIs to form unlabeled training samples by setting cropping stride to 400 pixels. The cropped tiles maintained 400$\times$400 size at all times. We split the labeled tiles into 3 categories: Normal/Benign, Cancer and Necrosis to perform 3-class semi-supervised classification tasks. Tile examples can be seen in Figure \ref{fig:institution_TCGA_tiles}(b). In summary, there are 84578, 180471, 7932 labeled training tiles, respectively, in the category of Normal/Benign, Cancer and Necrosis, as well as 19638, 79382, 1301 validation tiles, and 15323, 62565, 6168 test tiles respectively for each category. The number of unlabeled training tiles is 1,373,684. The TCGA ccRCC dataset is a publicly available dataset, and our annotations for the 150 WSIs are accessible at \url{github.com/BzhangURU/Paper_CLASS-M/tree/main/Section1_get_tiles_from_WSIs}, where we also provide our code to perform tile cropping. The tiles described above can be easily generated through our code by reading WSIs and annotations as inputs. 

\begin{table*}
\begin{center}
\caption{Performance of different classification models on Utah ccRCC dataset and TCGA ccRCC dataset. Mean accuracy and standard deviation on test set are calculated. The supervised setting was only trained on labeled samples. The encoders of self-supervised learning models including Barlow Twins, SwAV, MoCo v3, and ViT-DINO are pre-trained on both labeled and unlabeled samples. The results with the best mean accuracy are shown in bold. }
\label{ref:tab_test_result}
\begin{tabular}{|l|l|c|c|}
\hline
 &Models &  Test accuracy (Utah)& Test accuracy (TCGA)\\
\hline
\multirow{2}{*}{\shortstack[l]{{\bfseries Supervised} \\ (labeled images only)}}&ResNet  &  $88.85\pm 2.66\%$ & $72.11\pm 0.41\%$ \\ \cline{2-4}
 & ViT &$84.69\pm1.33\%$& $73.50\pm0.94\%$ \\ 
\specialrule{1.2pt}{0pt}{0pt} 
\multirow{4}{*}{\shortstack[l]{{\bfseries Self-supervised} \\ (pre-trained on labeled \\and unlabeled images)}} & Barlow Twins & $93.42\pm0.43\%$ & $77.42\pm4.92\%$ \\ \cline{2-4}
 & SwAV & $93.87\pm0.59\%$ & $82.17\pm0.05\%$ \\ \cline{2-4}
 & MoCo v3 & $93.91\pm0.60\%$ & $78.82\pm0.93\%$ \\ \cline{2-4}
 & ViT-DINO & $90.53\pm1.13\%$ & $79.76\pm2.15\%$ \\ \specialrule{1.2pt}{0pt}{0pt} 
\multirow{4}{*}{\shortstack[l]{{\bfseries Semi-supervised} \\ (trained on labeled \\and unlabeled images)}}&FixMatch & $91.58\pm 0.65\%$ & $83.34\pm 2.53\%$\\
\cline{2-4}
&MixMatch & $92.94\pm 1.54\%$ & $88.35\pm 1.39\%$\\
\cline{2-4}
&CLASS & $94.92 \pm 0.67\%$ & $83.06 \pm 0.47\%$\\
\cline{2-4}
&CLASS-M & $\mathbf{95.35 \pm 0.46\%}$ & $\mathbf{92.13 \pm 0.89\%}$\\
\hline
\end{tabular}
\end{center}
\end{table*}

\subsection{Adaptive stain separation}
In experiments, we maintained slide-by-slide level adaptive stain separation based on the following considerations: The stain styles vary across different WSIs due to different conditions. But within one WSI, most conditions, such as stain manufacturer, storage condition, and staining procedure, remain consistent. On the other hand, if adaptive stain separation is applied patch-by-patch, there are not enough pixel samples inside each patch to obtain robust stain separation results. For example, in Figure \ref{fig:institution_TCGA_tiles}(a), the top image in Necrosis category has very limited Hematoxylin portion, which makes stain separation less trustful at patch-by-patch level. 

The RGB to H/E conversion matrices for each WSI were computed before the CLASS-M model training. We provide our original code for adaptive stain separation based on algorithms in ~\cite{macenko2009method}. 
The link to our code can be found at \url{github.com/BzhangURU/Paper_CLASS-M/tree/main/Section2_get_stain_separation_matrices}

\subsection{Experiment settings}
For a fair comparison, we used ImageNet pre-trained ResNet18 model ~\cite{he2016deep} as an initialization for all Convolutional Neural Network (CNN) models in semi-supervised learning. We compared our CLASS-M model with other state-of-the-art semi-supervised learning models, including FixMatch ~\cite{sohn2020fixmatch} and MixMatch ~\cite{berthelot2019mixmatch}. 
In CLASS-M model training, we performed jittering of brightness, contrast and saturation on original RGB images, as well as image augmentations on H images and E images after stain separation, such as random rotation, random crop to $256\times256$, random flip and jittering of brightness. The image augmentations on H images and E images are independent, for example, they could have different rotation angles in random rotation. Since H images and E images are one-channel images, we summed the pre-trained weights on the first layer of ResNet18 as initialization weights. 

We used Root Mean Squared Propagation optimization method with a decaying learning rate. A batch size of 64 was chosen for all experiments. A balanced sampler was used to address the imbalance in the number of labels in the training set. In experiments with the Utah ccRCC dataset, we chose to have 32 labeled samples in each batch, with 8 samples for each category, while the remaining 32 samples were unlabeled. In experiments with the TCGA ccRCC dataset, we chose to have 33 labeled samples in each batch, with 11 samples for each category, and the remaining 31 samples would be unlabeled. In validation and test, we first calculated classification accuracies for each category, and then averaged them to get balanced validation accuracy and test accuracy. This approach ensures that each category holds equal importance, irrespective of the actual number of tiles for each category. The hyperparameters were fine-tuned for all models to get the best validation accuracy, then the test accuracy on that hyperparameter setting is reported as the final performance. We found that most hyperparameters didn't need to be changed between Utah ccRCC dataset and TCGA ccRCC dataset. We chose decaying learning rate with initial value set to $10^{-4}$. We set the number of augmentation times K to 2, temperature $T$ to 0.5, $\alpha$ to 2, margin $m$ in contrastive loss to 37 and unlabeled $L_2$ loss weight to 7.5 for both datasets. We set contrastive loss weight to 0.1 for experiments on the Utah ccRCC dataset, and contrastive loss weight to 0.001 for the TCGA ccRCC dataset. For each epoch, we ran 1000 iterations and checked validation accuracy once. The total epochs were large enough to ensure convergence. The training was stopped if the best validation accuracy was no longer updated for more than 100 epochs. Each experiment was repeated three times to obtain the average test accuracy and standard deviation.

The experiment platform is Python 3.7.11, Pytorch 1.9.0, torchvision 0.10.0, and CUDA 10.2. The GPUs we used are NVIDIA TITAN RTX. During CLASS-M model training, it usually took 13 GB GPU memory and 12 minutes for each training epoch. The code of our CLASS-M model is publicly available at \url{github.com/BzhangURU/Paper_CLASS-M/tree/main/Section3_CLASS-M}

\begin{table*}
\begin{center}
\caption{Ablation studies of CLASS-M models on Utah ccRCC dataset and TCGA ccRCC dataset. Mean accuracy and standard deviation on test set are calculated. The results with best mean accuracy are shown in bold. }
\label{ref:tab_ablation}
\begin{tabular}{|l|c|c|}
\hline
 Models &  Test accuracy (Utah)& Test accuracy (TCGA)\\
\hline
CLASS-M & $\mathbf{95.35 \pm 0.46\%}$ & $\mathbf{92.13 \pm 0.89\%}$\\
\hline
CLASS-M without contrastive loss & $90.92\pm 1.18\%$ & $89.70\pm 0.77\%$\\
\hline
CLASS-M without aug on RGB images & $94.41\pm 0.44\%$ & $91.21\pm 2.05\%$\\
\hline
CLASS-M without adaptive stain separation (use fixed stain separation) & $93.97\pm 0.40\%$ & $90.97\pm 1.86\%$\\
\hline
CLASS & $94.92 \pm 0.67\%$ & $83.06 \pm 0.47\%$\\
\hline
CLASS (use Red/Green channels as two views)& $90.75\pm 0.13\%$ & $81.14\pm 0.34\%$\\
\hline
CLASS (use Red/Blue channels as two views)& $89.06\pm 0.54\%$ & $80.14\pm 2.66\%$\\
\hline
CLASS (use Green/Blue channels as two views)& $83.43\pm 3.63\%$ & $80.25\pm 1.15\%$\\
\hline
CLASS without contrastive loss & $84.57\pm 2.40\%$ & $74.46\pm 2.03\%$\\
\hline
CLASS without aug on RGB images & $94.99\pm 0.58\%$ & $82.78\pm 1.10\%$\\
\hline
CLASS without adaptive stain separation (use fixed stain separation) & $92.43\pm 0.34\%$ & $79.89\pm 0.69\%$\\
\hline
CLASS without aug on RGB images and adaptive stain separation & $91.53\pm 1.52\%$ & $75.67\pm 1.76\%$\\
\hline

\end{tabular}
\end{center}
\end{table*}

\subsection{Comparison with baseline and other semi-supervised models}
We tested our CLASS-M model along with other state-of-the-art semi-supervised models, including FixMatch and MixMatch using the Utah ccRCC dataset and TCGA ccRCC datasets. CLASS-M without pseudo-labeling using MixUp, which we name CLASS, was tested as a reference. The baseline ResNet18 ~\cite{he2016deep} and Vision Transformer (ViT) ~\cite{dosovitskiy2020image} were also tested where only labeled samples were used in training. The ResNet18 and ViT models were initialized with weights pre-trained on ImageNet. 


As shown in Table~\ref{ref:tab_test_result}, CLASS-M model outperforms not only the supervised baselines, but also other state-of-the-art models by a large margin on both datasets. We achieved a test accuracy of $95.35 \pm 0.46\%$ on Utah ccRCC dataset, and $92.13 \pm 0.89\%$ on TCGA ccRCC dataset. 
The ResNet18 baseline results from Table~\ref{ref:tab_test_result} indicate that TCGA ccRCC dataset presents a harder challenge. 
One possible reason is that the samples from Utah ccRCC dataset have less variation in staining procedure, and the storage conditions are uniform. However, on TCGA ccRCC dataset, slides come from different institutions, leading to more variation in the quality of slides. Additionally, the Necrosis tiles on TCGA training set only come from 4 WSIs, which causes more difficulties in learning robust features for this class. As shown in the recall section of Table~\ref{ref:tab_recall_precision_F_TCGA} in Appenix B, pseudo-labeling with MixUp raised the Necrosis's test accuracy from 60.47\% to 86.65\% on the TCGA ccRCC dataset, demonstrating that MixUp is especially effective on classes with very limited number of samples. In conclusion, our CLASS-M model achieves the highest accuracy results on a variety of tasks. We provide recall, precision and F-score for each run, with details available in Appendix B. After obtaining the CLASS-M models, we ran the models on WSIs from the test set and generated prediction heatmaps to better visualize patch-level predictions. The results can be found in Appendix C.

\subsection{Comparison with self-supervised learning models}
To compare with self-supervised learning approaches, we used both labeled and unlabeled samples to pre-train self-supervised learning models on the Utah ccRCC dataset and TCGA ccRCC dataset separately. After pre-training, the encoder was frozen, and a fully connected classification network was appended. The final fully connected classification layer was trained only on labeled training samples. We selected several state-of-the-art self-supervised learning models for our experiments, including Barlow Twins ~\cite{zbontar2021barlow}, SwAV ~\cite{caron2020swav}, MoCo v3 ~\cite{chen2021empirical}, and ViT-DINO ~\cite{caron2021dino}, given their extensive use in histopathology~\cite{kang2023benchmarking}. As shown in Table~\ref{ref:tab_test_result}, our proposed models, CLASS and CLASS-M outperform those self-supervised learning models on ccRCC datasets. The recall, precision and F-score of the self-supervised learning models can also be found in Appendix B. The disadvantage of the semi-supervised approach compared to self-supervised learning is the loss of generality for the encoder which must be retrained for each new task. However, our results demonstrate that a well-formulated semi-supervised learning approach can have accuracy advantages over a self-supervised pre-training approach due to end-to-end learning with labeled and unlabeled samples simultaneously. Therefore, if accuracy is of paramount importance, semi-supervised learning may be preferred over self-supervised learning. 

\subsection{Ablation studies}
We conducted ablation studies to further analyze our model and identify the components contributing to the classification results. 
One of the critical parts of our model is the H/E contrastive loss which provides a regularization term and helps to generate representative features. As shown in Table~\ref{ref:tab_ablation}, removing the contrastive loss term led to a significant drop in the test accuracy. On the Utah ccRCC dataset, the test accuracy dropped from $95.35\%$ to $90.92\%$ for the CLASS-M model and from $94.92\%$ to $84.57\%$ for the CLASS model. On TCGA ccRCC dataset, the test accuracy declined from $92.13\%$ to $89.70\%$ for the CLASS-M model and from $83.06\%$ to $74.46\%$ for the CLASS model. 

Choosing Hematoxylin and Eosin channels as two views also plays a crucial role. Instead of choosing Hematoxylin and Eosin as two channels, we experimented with directly selecting two channels from Red, Green and Blue channels in RGB images to form two views in the CLASS model. As a result, we saw large reductions in test accuracies. This is expected due to the independence requirement in co-training which Hematoxylin and Eosin fulfill to a large extent, but Red/Green/Blue channels do not. In addition, splitting the original RGB images into Hematoxylin and Eosin channels is motivated from the pathology point of view and the true nature of H\&E images. 

The effect of adaptive stain separation was also verified. We used globally fixed stain separation matrix to replace adaptive stain separation. The CLASS-M model's test accuracy dropped from $95.35\%$ to $93.97\%$ on Utah ccRCC dataset and dropped from $92.13\%$ to $90.97\%$ on TCGA ccRCC dataset. We further did ablation study on the augmentations on original RGB images. The augmentations proved beneficial and improved classification accuracy from $94.41\%$ to $95.35\%$ on Utah ccRCC dataset and from $91.21\%$ to $92.13\%$ on TCGA ccRCC dataset for the CLASS-M model. We finally analyzed the role of pseudo-labeling using MixUp. Table~\ref{ref:tab_ablation} shows that, by comparing CLASS-M with CLASS, we saw a substantial improvement by adding MixUp augmentation, especially on the TCGA ccRCC dataset. 

\section{Conclusion}
In this paper, we proposed a novel semi-supervised model named CLASS-M for histopathological image classification by introducing adaptive stain separation-based contrastive learning and adopting pseudo-labeling using MixUp. We provided the newly annotated Utah ccRCC dataset and TCGA ccRCC dataset. Experiments have shown that our CLASS-M model consistently reached the best classification results compared to other state-of-the-art models. Our model demonstrates the capability to perform accurate patch-level classification at various resolutions with only rough annotations on approximately 30 WSIs in training. We demonstrated that CLASS-M outperforms state-of-the-art general semi-supervised computer vision models. The code for our model is also publicly available. 

The advantage of semi-supervised learning is the end-to-end training despite the sacrifice of conveniency compared to self-supervised learning. Self-supervised learning freezes the pre-trained encoders in final training with labeled data, which lacks the flexibility to fine-tune the whole model. However, unfreezing the encoders in final training leads to overfitting as there is no longer access to the large unlabeled dataset, especially when labeled data is limited. We demonstrated that our semi-supervised model is able to outperform self-supervised learning models. 

Future work may involve addressing the challenge of handling noisy labels. In our datasets, there is a small portion of tiles that only contain blood vessels, which makes labels inaccurate. A model with more capabilities to tolerate noisy labels may be an interesting road to explore. Our method can also be readily applied to other types of histopathological images, such as immunohistochemistry (IHC) stained images. 

\section*{CRediT author statement}

\textbf{Bodong Zhang}: Conceptualization, Methodology, Software, Validation, Formal analysis, Investigation, Data Curation, Writing - Original Draft, Visualization. 
\textbf{Hamid Manoochehri}: Software, Formal analysis, Visualization. 
\textbf{Man Minh Ho}: Software, Validation, Investigation. 
\textbf{Fahimeh Fooladgar}: Software, Validation, Investigation. 
\textbf{Yosep Chong}: Writing - Review \& Editing. 
\textbf{Beatrice S. Knudsen}: Conceptualization, Resources, Data Curation, Writing - Review \& Editing, Supervision, Project administration, Funding acquisition. 
\textbf{Deepika Sirohi}: Resources, Data Curation. 
\textbf{Tolga Tasdizen}: Conceptualization, Resources, Writing - Review \& Editing, Supervision, Project administration, Funding acquisition.

\section*{Availability of research data/code}
Our related research data and code is publicly available at \url{github.com/BzhangURU/Paper_CLASS-M/tree/main}

\section*{Acknowledgments}
The work was funded in part by NIH 1R21CA277381. We also acknowledge the support of the Computational Oncology Research Initiative (CORI) at the Huntsman Cancer Institute, ARUP Laboratories, and the Department of Pathology at the University of Utah. The project utilized H\&E images from the KIRK cohort of the TCGA Research Network: https://www.cancer.gov/tcga

\section*{Appendix A. Detailed algorithm of Adaptive Stain Seperation}
First, the RGB pixel values from slide images are transformed into Optical Density (OD) space, where stain components are formed by linear combinations. The corresponding Beer-Lambert law ~\cite{beer1852bestimmung,lambert1760photometria} for this operation is below:
\begin{equation}
    OD_{C} = \log_{10} \frac{I_{0,C}}{I_{C}}
\end{equation}
The letter C represents a specific channel such as Red, Green, or Blue channel. The value $I_{0,C}$ refers to the intensity of light before passing through the specimen, which is background intensity, and the value $I_{C}$ is the intensity of light after passing through the specimen. We can treat the $OD$ value as a measurement of the absorption of light on different channels. Zero values in all channels on $OD$ space correspond to cases of pure white background where there is no tissue absorbing light on that pixel. For each RGB pixel, we use a $3\times1$ vector $V_{OD}=(OD_{R}, OD_{G}, OD_{B})^{\top}$ to show the values on OD space.

\begin{figure*}
    \centering
    \includegraphics[scale=.45]{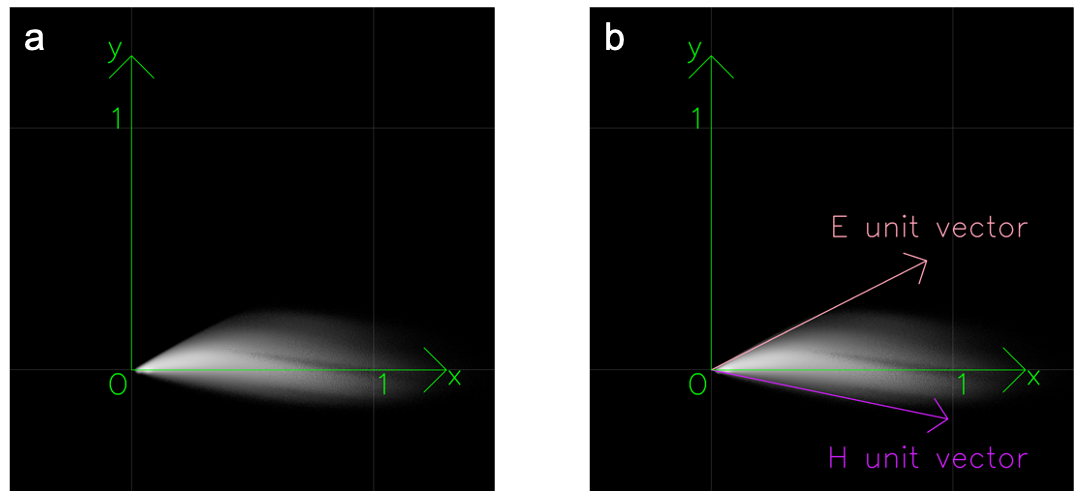}
    \caption{Mapping pixels onto the 2D OD space}
    \label{fig:2D_OD}
\end{figure*}

Second, dimensionality reduction is applied. In an ideal situation of H\&E staining, $V_{OD}$ is a linear combination of fixed Hematoxylin and Eosin unit vectors: $V_{OD}=\alpha_{H} V_{H}+\alpha_{E} V_{E}$, where $V_H$ and $V_E$ are $3\times1$ unit vectors on 3D OD space. Therefore, all OD vectors should approximately fall into the same 2D plane for all pixels in one image. We use Principal Component Analysis (PCA) on 3D OD space to get a 2D plane formed by two covariance matrix's eigenvectors with the largest two eigenvalues. Figure \ref{fig:2D_OD}.a is an example of the distribution of OD vectors that map onto the 2D plane. The brightness shows the density of pixels. The x-axis is the direction of the eigenvector with the largest eigenvalue. The y-axis is the direction of the eigenvector with the second-largest eigenvalue. The third dimension can be treated as a residual part and abandoned. The x and y axes are orthogonal because the eigenvectors of the covariance matrix are orthogonal to each other. By applying the PCA method, the 2D plane we find is more informative, ensuring that the separated H and E images contain critical information and details.

The third step is to find stain vectors $V_H$ and $V_E$. ~\cite{macenko2009method} assumes every pixel sample in OD space must exist between the two stain vectors, considering the fact that each component should be non-negative. However, noise can't be fully avoided. As a result, we allow 1\% of samples to be outside of $V_H$ and $V_E$ respectively. Moreover, pixel samples that are within a distance of 0.1 from origin are not taken into account as they contain less stain and are more easily influenced by noise. An example of $V_H$ and $V_E$ is shown in Figure \ref{fig:2D_OD}.b. The positive directions of x-axis and y-axis were carefully chosen such that H unit vector falls into quadrant IV and E unit vector falls into quadrant I. 

After acquiring unit vectors $V_H$ and $V_E$, we are able to reconstruct the transformation matrix. In the previous PCA process, we computed three unit eigenvectors $V_x$, $V_y$, $V_{Residual}$, which are orthogonal to each other and ordered by their eigenvalues from maximum to minimum. The $V_H$ and $V_E$ on x-y 2D OD space can be written as $V_H = \cos \theta_{H} V_{x} + \sin \theta_{H} V_{y}$, $V_{E} = \cos \theta_{E} V_{x} + \sin \theta_{E} V_{y}$. 
The transformation matrix from H, E, Residual to RGB on OD space is found as
\begin{equation}
  Mat_{HERes\rightarrow RGB\,on\,OD} = [V_{H}, V_{E}, V_{Residual}]  
\end{equation}
Therefore:
\begin{equation}
\begin{split}
&(OD_{R}, OD_{G}, OD_{B})^\mathsf{T} = \\
&Mat_{HERes \rightarrow RGB\,on\,OD} \times (\alpha_{H} , \alpha_{E} , \alpha_{Residual})^\mathsf{T}
\end{split}
\end{equation}

If we define $Mat_{RGB \rightarrow HERes\,on\,OD}$ as the inverse of $Mat_{HERes \rightarrow RGB\,on\,OD}$, then:
\begin{equation}
\begin{split}
&(\alpha_H , \alpha_E , \alpha_{Residual})^\mathsf{T} = \\
&Mat_{RGB \rightarrow HERes\,on\,OD} \times (OD_R , OD_G , OD_B)^\mathsf{T}
\end{split}
\end{equation}

If we apply this formula to all pixels in an image, then the $\alpha_H$ forms the Hematoxylin image and $\alpha_E$ forms the Eosin image.

Finally, to further normalize Hematoxylin images and Eosin images, the 99th percentile of intensity values is used as an approximation of maximum value. We normalize it to 0.5 and clip to 1.0 for any numbers larger than 1.0 after normalization. 

The main advantages of this stain separation method are simplicity and efficiency. We are able to perform it without the need of model training or complex calculations. Moreover, finding optimum H vectors and E vectors helps to handle stain variations. The final normalization based on the calculated maximum value further mitigates the effect of variations on the brightness and strength of stains.  

\begin{table*}
\begin{center}
\caption{Recall, precision and F-score of different classification models on Utah ccRCC dataset. Mean and standard deviation on test set are calculated based on 3 runs. Blue numbers show fully-supervised learning results. Green numbers show self-supervised learning results. Black numbers show semi-supervised learning results.}
\label{ref:tab_recall_precision_F_our_institution}
\begin{tabular}{|c|l|c|c|c|c|c|}
\hline
 &Models &  Normal Tissue & Low Risk Cancer & High Risk Cancer & Necrosis &average (all classes)\\
\hline
\multirow{10}{*}{Recall}&ResNet  &\textcolor{myblue}{$0.9308\pm0.0358$}&\textcolor{myblue}{$0.9186\pm0.0254$}&\textcolor{myblue}{$0.9717\pm0.0384$}&\textcolor{myblue}{$0.7330\pm0.1172$}&  \textcolor{myblue}{$0.8885\pm0.0266$}  \\ \cline{2-7}
 & ViT &\textcolor{myblue}{$0.8906\pm0.0138$}&\textcolor{myblue}{$0.9571\pm0.0179$}&\textcolor{myblue}{$0.7112\pm0.0351$}&\textcolor{myblue}{$0.8286\pm0.0278$}&\textcolor{myblue}{$0.8469\pm0.0133$}\\ \cline{2-7}
 & BT &\textcolor{mygreen}{$0.9685\pm0.0030$ }&\textcolor{mygreen}{$0.9693\pm0.0026$ }&\textcolor{mygreen}{$0.9169\pm0.0083$ }&\textcolor{mygreen}{$0.8820\pm0.0097$ }&\textcolor{mygreen}{$0.9342\pm0.0043$} \\ \cline{2-7}
 & SwAV &\textcolor{mygreen}{$0.9413\pm0.0049$} &\textcolor{mygreen}{$0.9889\pm0.0010$} &\textcolor{mygreen}{$0.9674\pm0.0039$} &\textcolor{mygreen}{$0.8571\pm0.0242$} &\textcolor{mygreen}{$0.9387\pm0.0059$} \\ \cline{2-7}
 & MoCo v3 &\textcolor{mygreen}{$0.9631\pm0.0068$} &\textcolor{mygreen}{$0.9755\pm0.0010$} &\textcolor{mygreen}{$0.9306\pm0.0089$} &\textcolor{mygreen}{$0.8874\pm0.0082$} &\textcolor{mygreen}{$0.9391\pm0.0060$} \\ \cline{2-7}
 & ViT-DINO &\textcolor{mygreen}{$0.9017\pm0.0230$} &\textcolor{mygreen}{$0.9933\pm0.0033$} &\textcolor{mygreen}{$0.8946\pm0.0567$} &\textcolor{mygreen}{$0.8315\pm0.0357$} &\textcolor{mygreen}{$0.9053\pm0.0113$}\\ \cline{2-7}
&FixMatch &$0.8322\pm0.0759$&$0.9693\pm0.0459$&$0.9563\pm0.0602$&$0.9051\pm0.0558$& $0.9158\pm 0.0065$ \\ \cline{2-7}
&MixMatch &$0.9004\pm0.0261$&$0.9047\pm0.0420$&$0.9588\pm0.0044$&$0.9538\pm0.0092$& $0.9294\pm0.0154$ \\ \cline{2-7}
&CLASS &$0.9753\pm0.0061$&$0.9660\pm0.0092$&$0.9563\pm0.0051$&$0.8994\pm0.0260$& $0.9492\pm0.0067$ \\ \cline{2-7}
&CLASS-M &$0.9252\pm0.0194$&$0.9576\pm0.0174$&$0.9880\pm0.0065$&$0.9430\pm0.0051$& $0.9535\pm0.0046$ \\ \specialrule{1.5pt}{0pt}{0pt} 
\multirow{10}{*}{Precision}
&ResNet  &\textcolor{myblue}{$0.9702\pm0.0148$}&\textcolor{myblue}{$0.8665\pm0.1087$}&\textcolor{myblue}{$0.4963\pm0.1317$}&\textcolor{myblue}{$0.8930\pm0.0493$}&\textcolor{myblue}{$0.8065\pm0.0493$}    \\ \cline{2-7}
 & ViT &\textcolor{myblue}{$0.9832\pm0.0034$}&\textcolor{myblue}{$0.5781\pm0.0642$}&\textcolor{myblue}{$0.3941\pm0.0429$}&\textcolor{myblue}{$0.8649\pm0.0547$}&\textcolor{myblue}{$0.7051\pm0.0132$}\\ \cline{2-7}
& BT &\textcolor{mygreen}{$0.9924\pm0.0008$ }&\textcolor{mygreen}{$0.9021\pm0.0100$ }&\textcolor{mygreen}{$0.7107\pm0.0134$ }&\textcolor{mygreen}{$0.8666\pm0.0277$ }&\textcolor{mygreen}{$0.8679\pm0.0076$} \\ \cline{2-7}
 & SwAV &\textcolor{mygreen}{$0.9855\pm0.0031$} &\textcolor{mygreen}{$0.9589\pm0.0018$} &\textcolor{mygreen}{$0.6785\pm0.0070$} &\textcolor{mygreen}{$0.7442\pm0.0212$} &\textcolor{mygreen}{$0.8418\pm0.0029$} \\ \cline{2-7}
 & MoCo v3 &\textcolor{mygreen}{$0.9860\pm0.0010$} &\textcolor{mygreen}{$0.9149\pm0.0132$} &\textcolor{mygreen}{$0.8037\pm0.0276$} &\textcolor{mygreen}{$0.8271\pm0.0279$} &\textcolor{mygreen}{$0.8829\pm0.0171$} \\ \cline{2-7}
 & ViT-DINO &\textcolor{mygreen}{$0.9803\pm0.0023$} &\textcolor{mygreen}{$0.7384\pm0.0237$} &\textcolor{mygreen}{$0.6783\pm0.1003$} &\textcolor{mygreen}{$0.6655\pm0.0961$} &\textcolor{mygreen}{$0.7657\pm0.0358$} \\ \cline{2-7}
&FixMatch &$0.9913\pm0.0059$&$0.6529\pm0.1011$&$0.3679\pm0.0784$&$0.7837\pm0.1874$& $0.6990\pm 0.0776$ \\ \cline{2-7}
&MixMatch &$0.9945\pm0.0015$&$0.8400\pm0.1067$&$0.4072\pm0.0615$&$0.8660\pm0.0518$& $0.7770\pm0.0401$ \\ \cline{2-7}
&CLASS &$0.9918\pm0.0026$&$0.9165\pm0.0103$&$0.7164\pm0.0411$&$0.9439\pm0.0119$& $0.8921\pm0.0111$ \\ \cline{2-7}
&CLASS-M &$0.9980\pm0.0005$&$0.9263\pm0.0102$&$0.7316\pm0.0048$&$0.6760\pm0.0666$& $0.8330\pm0.0192$ \\ \specialrule{1.5pt}{0pt}{0pt} 
\multirow{10}{*}{F-score}
&ResNet  &\textcolor{myblue}{$0.9497\pm0.0137$}&\textcolor{myblue}{$0.8903\pm0.0706$}&\textcolor{myblue}{$0.6483\pm0.1052$}&\textcolor{myblue}{$0.8005\pm0.0641$}&\textcolor{myblue}{$0.8222\pm0.0256$}  \\ \cline{2-7}
& ViT &\textcolor{myblue}{$0.9346\pm0.0063$}&\textcolor{myblue}{$0.7192\pm0.0495$}&\textcolor{myblue}{$0.5065\pm0.0400$}&\textcolor{myblue}{$0.8452\pm0.0152$}&\textcolor{myblue}{$0.7513\pm0.0156$}\\ \cline{2-7}
 & BT &\textcolor{mygreen}{$0.9803\pm0.0018$ }&\textcolor{mygreen}{$0.9345\pm0.0044$ }&\textcolor{mygreen}{$0.8007\pm0.0079$ }&\textcolor{mygreen}{$0.8741\pm0.0167$ }&\textcolor{mygreen}{$0.8974\pm0.0060$} \\ \cline{2-7}
 & SwAV &\textcolor{mygreen}{$0.9628\pm0.0011$} &\textcolor{mygreen}{$0.9737\pm0.0014$} &\textcolor{mygreen}{$0.7976\pm0.0062$} &\textcolor{mygreen}{$0.7963\pm0.0020$} &\textcolor{mygreen}{$0.8826\pm0.0012$} \\ \cline{2-7}
 & MoCo v3 &\textcolor{mygreen}{$0.9744\pm0.0039$} &\textcolor{mygreen}{$0.9442\pm0.0074$} &\textcolor{mygreen}{$0.8624\pm0.0197$} &\textcolor{mygreen}{$0.8561\pm0.0180$} &\textcolor{mygreen}{$0.9093\pm0.0122$} \\ \cline{2-7}
 & ViT-DINO &\textcolor{mygreen}{$0.9393\pm0.0132$} &\textcolor{mygreen}{$0.8469\pm0.0146$} &\textcolor{mygreen}{$0.7662\pm0.0428$} &\textcolor{mygreen}{$0.7381\pm0.0721$} &\textcolor{mygreen}{$0.8226\pm0.0235$} \\ \cline{2-7}

&FixMatch &$0.9035\pm0.0415$&$0.7785\pm0.0834$&$0.5263\pm0.0777$&$0.8273\pm0.0878$& $0.7589\pm0.0572$ \\ \cline{2-7}
&MixMatch &$0.9451\pm0.0137$&$0.8692\pm0.0697$&$0.5699\pm0.0600$&$0.9072\pm0.0281$& $0.8228\pm0.0357$ \\ \cline{2-7}
&CLASS &$0.9835\pm0.0019$&$0.9406\pm0.0093$&$0.8187\pm0.0267$&$0.9208\pm0.0080$& $0.9159\pm0.0044$ \\ \cline{2-7}
&CLASS-M &$0.9602\pm0.0106$&$0.9416\pm0.0037$&$0.8407\pm0.0050$&$0.7864\pm0.0479$& $0.8822\pm0.0150$ \\ \hline
\end{tabular}
\end{center}
\end{table*}

\begin{table*}
\begin{center}
\caption{Recall, precision and F-score of different classification models on TCGA ccRCC dataset. Mean and standard deviation on test set are calculated based on 3 runs. Blue numbers show fully-supervised learning results. Green numbers show self-supervised learning results. Black numbers show semi-supervised learning results.}
\label{ref:tab_recall_precision_F_TCGA}
\begin{tabular}{|c|l|c|c|c|c|}
\hline
 &Models &  Normal Tissue & Cancer & Necrosis & average (all classes) \\
\hline
\multirow{12}{*}{Recall}
&ResNet  &\textcolor{myblue}{$0.7136\pm0.0191$}&\textcolor{myblue}{$0.9032\pm0.0105$}&\textcolor{myblue}{$0.5466\pm0.0172$}&\textcolor{myblue}{$0.7211\pm0.0041$}    \\ \cline{2-6}
& ViT &\textcolor{myblue}{$0.7502\pm0.0164$ }&\textcolor{myblue}{$0.8986\pm0.0130$ }&\textcolor{myblue}{$0.5563\pm0.0242$ }&\textcolor{myblue}{$0.7350\pm0.0094$} \\ \cline{2-6}
 & BT &\textcolor{mygreen}{$0.8390\pm0.0322$} &\textcolor{mygreen}{$0.9003\pm0.0145$} &\textcolor{mygreen}{$0.5833\pm0.1345$} &\textcolor{mygreen}{$0.7742\pm0.0492$} \\ \cline{2-6}
 & SwAV &\textcolor{mygreen}{$0.9542\pm0.0029$} &\textcolor{mygreen}{$0.9714\pm0.0011$} &\textcolor{mygreen}{$0.5393\pm0.0029$} &\textcolor{mygreen}{$0.8217\pm0.0005$} \\ \cline{2-6}
 & MoCo v3 &\textcolor{mygreen}{$0.8479\pm0.0329$} &\textcolor{mygreen}{$0.9559\pm0.0078$} &\textcolor{mygreen}{$0.5609\pm0.0030$} &\textcolor{mygreen}{$0.7882\pm0.0093$} \\ \cline{2-6}
 & ViT-DINO &\textcolor{mygreen}{$0.9222\pm0.0511$} &\textcolor{mygreen}{$0.9857\pm0.0047$} &\textcolor{mygreen}{$0.4850\pm0.0577$} &\textcolor{mygreen}{$0.7976\pm0.0215$} \\ \cline{2-6}

&FixMatch &$0.8582\pm0.0548$&$0.9565\pm0.0165$&$0.6855\pm0.0664$& $0.8334\pm0.0253$ \\ \cline{2-6}
&MixMatch &$0.9198\pm0.0334$&$0.8910\pm0.0250$&$0.8397\pm0.0156$& $0.8835\pm0.0139$ \\ \cline{2-6}
&CLASS &$0.9291\pm0.0377$&$0.9580\pm0.0212$&$0.6047\pm0.0260$& $0.8306\pm0.0047$ \\ \cline{2-6}
&CLASS-M &$0.9477\pm0.0173$&$0.9496\pm0.0060$&$0.8665\pm0.0106$& $0.9213\pm0.0089$ \\ \specialrule{1.5pt}{0pt}{0pt} 
\multirow{12}{*}{Precision}
&ResNet  &\textcolor{myblue}{$0.6168\pm0.0214$}&\textcolor{myblue}{$0.9130\pm0.0048$}&\textcolor{myblue}{$0.7622\pm0.0128$}&\textcolor{myblue}{$0.7640\pm0.0062$}    \\ \cline{2-6}
& ViT &\textcolor{myblue}{$0.6281\pm0.0302$ }&\textcolor{myblue}{$0.9166\pm0.0035$ }&\textcolor{myblue}{$0.7879\pm0.0902$ }&\textcolor{myblue}{$0.7776\pm0.0385$} \\ \cline{2-6}
 & BT &\textcolor{mygreen}{$0.7161\pm0.0457$} &\textcolor{mygreen}{$0.9398\pm0.0256$} &\textcolor{mygreen}{$0.6184\pm0.1002$} &\textcolor{mygreen}{$0.7581\pm0.0162$} \\ \cline{2-6}
 & SwAV &\textcolor{mygreen}{$0.8597\pm0.0039$} &\textcolor{mygreen}{$0.9581\pm0.0011$} &\textcolor{mygreen}{$0.9211\pm0.0055$} &\textcolor{mygreen}{$0.9130\pm0.0025$} \\ \cline{2-6}
 & MoCo v3 &\textcolor{mygreen}{$0.7902\pm0.0241$} &\textcolor{mygreen}{$0.9404\pm0.0100$} &\textcolor{mygreen}{$0.8681\pm0.0102$} &\textcolor{mygreen}{$0.8662\pm0.0082$} \\ \cline{2-6}
 & ViT-DINO &\textcolor{mygreen}{$0.9420\pm0.0244$} &\textcolor{mygreen}{$0.9374\pm0.0100$} &\textcolor{mygreen}{$0.9210\pm0.0125$} &\textcolor{mygreen}{$0.9335\pm0.0068$} \\ \cline{2-6}
&FixMatch &$0.8651\pm0.0677$&$0.9530\pm0.0151$&$0.7338\pm0.1257$& $0.8506\pm0.0224$ \\ \cline{2-6}
&MixMatch &$0.7505\pm0.0380$&$0.9939\pm0.0031$&$0.5674\pm0.0380$& $0.7706\pm0.0235$ \\ \cline{2-6}
&CLASS &$0.8727\pm0.0472$&$0.9636\pm0.0108$&$0.6986\pm0.1187$& $0.8449\pm0.0452$ \\ \cline{2-6}
&CLASS-M &$0.8592\pm0.0249$&$0.9913\pm0.0037$&$0.7420\pm0.0295$& $0.8641\pm0.0130$ \\ \specialrule{1.5pt}{0pt}{0pt} 
\multirow{12}{*}{F-score}
&ResNet  &\textcolor{myblue}{$0.6616\pm0.0183$}&\textcolor{myblue}{$0.9081\pm0.0076$}&\textcolor{myblue}{$0.6365\pm0.0117$}&\textcolor{myblue}{$0.7354\pm0.0051$}    \\ \cline{2-6}
& ViT &\textcolor{myblue}{$0.6836\pm0.0222$ }&\textcolor{myblue}{$0.9075\pm0.0056$ }&\textcolor{myblue}{$0.6503\pm0.0308$ }&\textcolor{myblue}{$0.7471\pm0.0192$} \\ \cline{2-6}
 & BT &\textcolor{mygreen}{$0.7718\pm0.0266$} &\textcolor{mygreen}{$0.9194\pm0.0098$} &\textcolor{mygreen}{$0.5849\pm0.0325$} &\textcolor{mygreen}{$0.7587\pm0.0195$} \\ \cline{2-6}
 & SwAV &\textcolor{mygreen}{$0.9045\pm0.0009$} &\textcolor{mygreen}{$0.9647\pm0.0006$} &\textcolor{mygreen}{$0.6803\pm0.0032$} &\textcolor{mygreen}{$0.8498\pm0.0014$} \\ \cline{2-6}
 & MoCo v3 &\textcolor{mygreen}{$0.8174\pm0.0050$} &\textcolor{mygreen}{$0.9481\pm0.0012$} &\textcolor{mygreen}{$0.6814\pm0.0021$} &\textcolor{mygreen}{$0.8156\pm0.0022$} \\ \cline{2-6}
 & ViT-DINO &\textcolor{mygreen}{$0.9314\pm0.0264$} &\textcolor{mygreen}{$0.9609\pm0.0058$} &\textcolor{mygreen}{$0.6337\pm0.0463$} &\textcolor{mygreen}{$0.8420\pm0.0200$} \\ \cline{2-6}
&FixMatch &$0.8588\pm0.0120$&$0.9546\pm0.0062$&$0.7016\pm0.0439$& $0.8383\pm0.0104$ \\ \cline{2-6}
&MixMatch &$0.8260\pm0.0273$&$0.9395\pm0.0132$&$0.6767\pm0.0282$& $0.8141\pm0.0209$ \\ \cline{2-6}
&CLASS &$0.8987\pm0.0084$&$0.9606\pm0.0061$&$0.6435\pm0.0429$& $0.8343\pm0.0186$ \\ \cline{2-6}
&CLASS-M &$0.9011\pm0.0150$&$0.9700\pm0.0017$&$0.7993\pm0.0211$& $0.8901\pm0.0109$ \\ \hline
\end{tabular}
\end{center}
\end{table*}

\section*{Appendix B. More details on experiment results.}
Please check Table~\ref{ref:tab_recall_precision_F_our_institution} and Table~\ref{ref:tab_recall_precision_F_TCGA} for recall, precision and F-score of each model on Utah ccRCC dataset and TCGA ccRCC dataset. 

\section*{Appendix C. Visualization of our CLASS-M model predictions.}
\begin{figure*}
    \centering
    \includegraphics[width=1.0\textwidth]{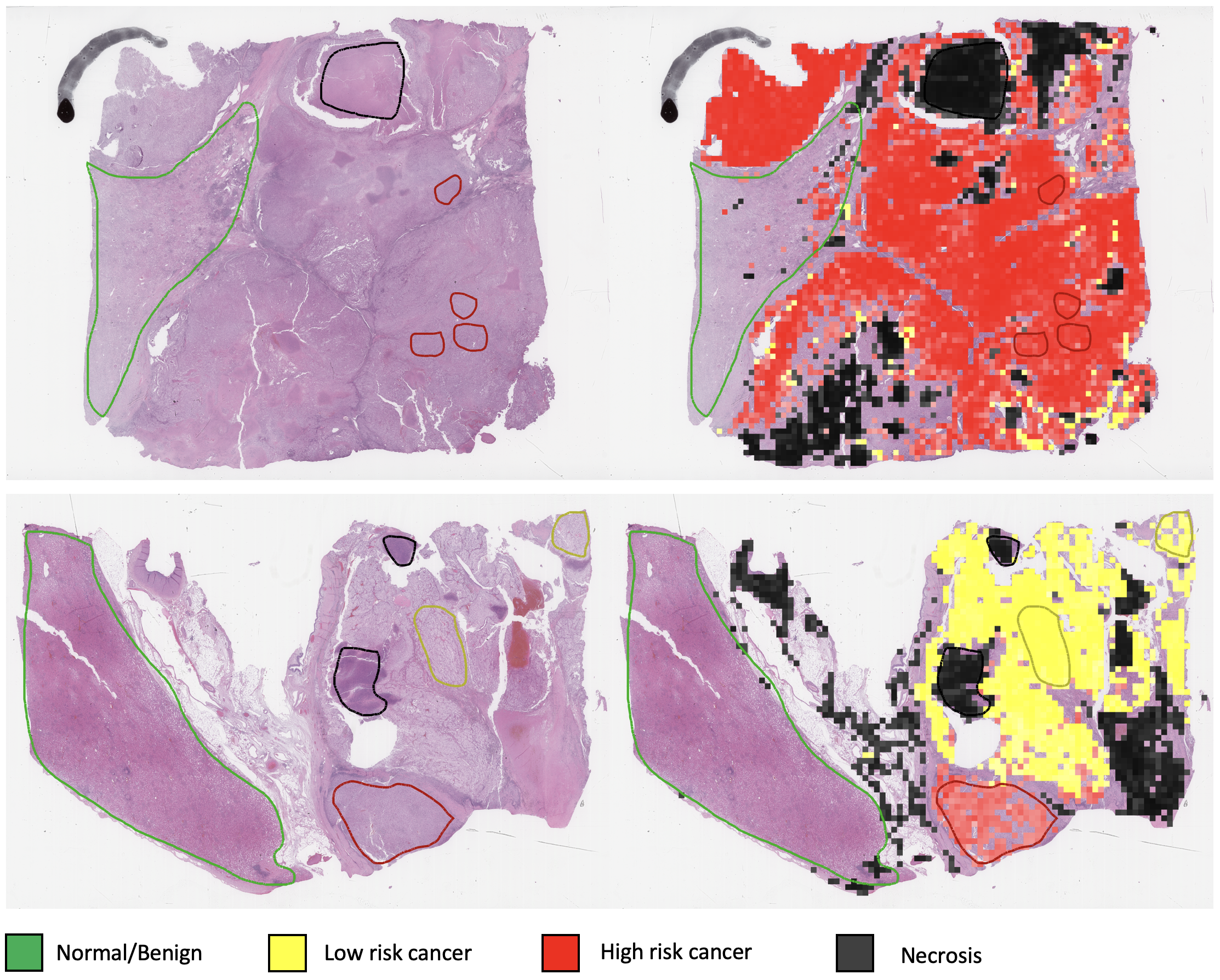}
    \caption{Visualization of our CLASS-M model's predictions on Utah ccRCC WSIs from test set. The polygons show the annotations on WSIs. (Green: Normal/Benign, Yellow: Low risk cancer, Red: High risk cancer, Dark Grey: Necrosis) Each polygon labeled as low/high risk cancer marks the region of a certain cancer growth pattern. The heatmaps show the model's predictions on all foreground tiles when the predictions are not Normal/Benign. The strength of color shows the confidence of prediction. }
    \label{fig:pred_heatmap_Utah}
\end{figure*}

\begin{figure*}
    \centering
    \includegraphics[width=1.0\textwidth]{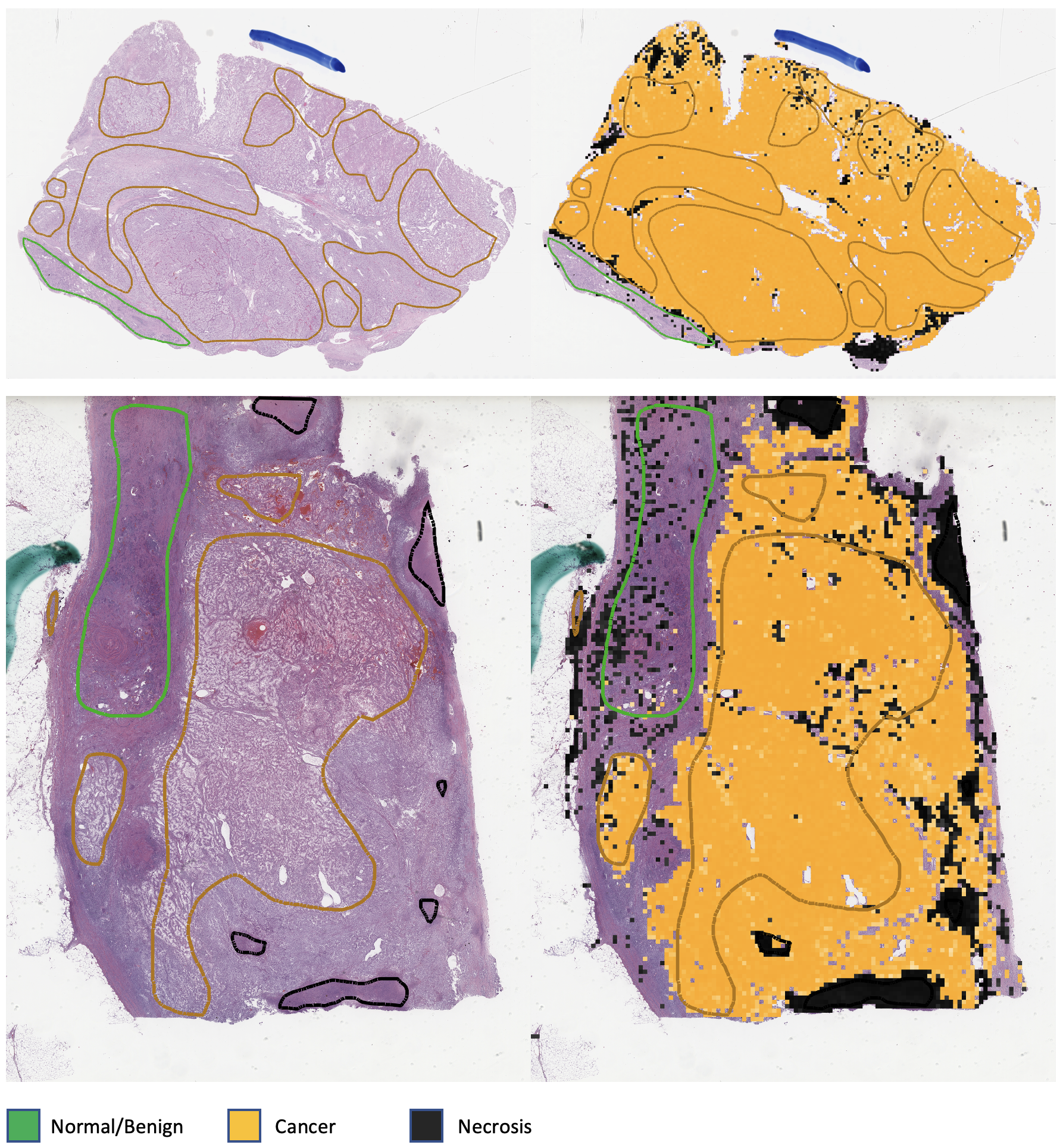}
    \caption{Visualization of our CLASS-M model's predictions on TCGA ccRCC WSIs from test set. The polygons show the annotations on WSIs. (Green: Normal/Benign, Orange: Cancer, Dark Grey: Necrosis) Each polygon labeled as cancer marks the region of a certain cancer growth pattern. The heatmaps show the model's predictions on all foreground tiles when the predictions are not Normal/Benign. The strength of color shows the confidence of prediction. }
    \label{fig:pred_heatmap_TCGA}
\end{figure*}

Please check Figure \ref{fig:pred_heatmap_Utah} and Figure \ref{fig:pred_heatmap_TCGA} for visualization of our CLASS-M model's patch-level predictions on WSIs from test set for both datasets.

\bibliographystyle{splncs04}
\bibliography{refs}

\end{multicols}

\end{document}